%% file: main.tex
\pdfoutput=1

\documentclass[11pt]{article}

\usepackage[final]{acl}

\usepackage{times}
\usepackage{latexsym}
\usepackage{hyperref}
\usepackage{url}
\usepackage{graphicx}
\usepackage{amsmath}
\usepackage{amsfonts}
\usepackage{amsthm}
\usepackage{bbm}
\usepackage{booktabs}
\usepackage{caption}
\usepackage{multirow}
\usepackage{subfigure}
\usepackage{subcaption}
\usepackage{pgfplots}
\usepackage{xspace}
\usepackage{enumitem}
\usepackage{wrapfig}
\usepackage{bbding}
\usepackage{colortbl}
\usepackage{amsthm}
\usepackage[ruled,vlined,linesnumbered]{algorithm2e}
\usepackage{sidecap}

\definecolor{mycolor_green}{HTML}{D5E8D4}
\definecolor{mycolor_red}{HTML}{F8CECC}
\definecolor{mycolor_blue}{HTML}{DAE8FC}
\definecolor{mycolor_orange}{HTML}{FFE6CC}

\DeclareUnicodeCharacter{1EC5}{\^e\u{}}
\DeclareUnicodeCharacter{1EAD}{\d{a}}

\newcommand{\mycolorbox}[2]{%
  \begingroup
  \setlength{\fboxsep}{0pt}
  \colorbox{#1}{#2}
  \endgroup
}

\newcommand*\samethanks[1][\value{footnote}]{\footnotemark[#1]}

\newcommand{\modelEmoji}{\includegraphics[height=1.4em,trim=0 .8em 0 0]{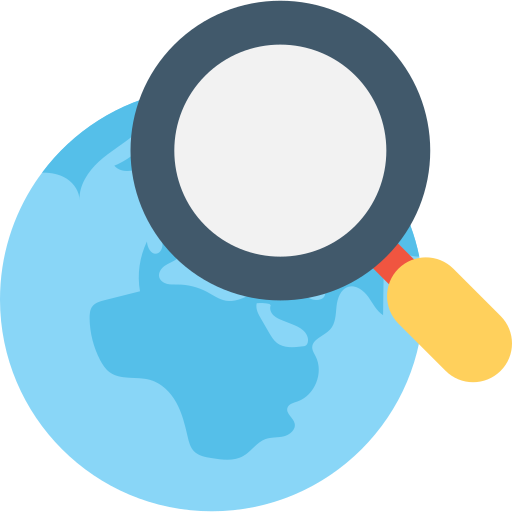}}
\newcommand{\modelWithEmoji}{\modelEmoji \textsc{Lens}\xspace}

\usepackage[T1]{fontenc}

\usepackage[utf8]{inputenc}

\usepackage{microtype}

\usepackage{inconsolata}

\usepackage{graphicx}

\title{\modelWithEmoji: Rethinking Multilingual Enhancement \\ for Large Language Models}

\author{%
  Weixiang Zhao\textsuperscript{1}, Yulin Hu\textsuperscript{1}\thanks{\ \ Equal contribution} \ Jiahe Guo\textsuperscript{1}\samethanks \ Xingyu Sui\textsuperscript{1}\samethanks, Tongtong Wu\textsuperscript{2} \\ \textbf{Yang Deng}\textsuperscript{3}
  \textbf{Yanyan Zhao\textsuperscript{1}},  \textbf{Bing Qin\textsuperscript{1}}, \textbf{Wanxiang Che\textsuperscript{1}}, \textbf{Ting Liu\textsuperscript{1}}\\
  \textsuperscript{1}Harbin Institute of Technology, \textsuperscript{2}Monash University\\
  \textsuperscript{3}Singapore Management University \\
  \texttt{\{wxzhao,yyzhao,qinb,car,tliu\}@ir.hit.edu.cn} \\
}

\begin{document}
\maketitle
\begin{abstract}
As global demand for multilingual large language models (LLMs) grows, most LLMs still remain overly focused on English, leading to the limited access to advanced AI for non-English speakers. Current methods to enhance multilingual capabilities largely rely on data-driven post-training techniques, such as multilingual instruction tuning or continual pre-training. However, these approaches exhibit significant limitations, including high resource cost, exacerbation of off-target issue and catastrophic forgetting of central language abilities. To this end, we propose \textsc{Lens}, a novel approach that enhances multilingual capabilities by leveraging LLMs’ internal language representation spaces. \textsc{Lens} operates on two subspaces: the language-agnostic subspace, where it aligns target languages with the central language to inherit strong semantic representations, and the language-specific subspace, where it separates target and central languages to preserve linguistic specificity. Experiments on three English-centric LLMs show that \textsc{Lens} significantly improves multilingual performance while maintaining the model’s English proficiency, achieving better results with less computational cost compared to existing post-training approaches. Our code is available at \url{https://github.com/circle-hit/Lens}.
\end{abstract}

\section{Introduction}

In an increasingly interconnected world, large language models (LLMs) are expected to cater to a diverse range of users across various linguistic backgrounds \citep{ouyang2023shifted,zhaowildchat,zhenglmsys,qin2024multilingual}. However, most state-of-the-art LLMs remain heavily English-centric \citep{brown2020language,touvron2023llama1,touvron2023llama2,jiang2023mistral,llama3modelcard}, performing far better in English than other languages, thereby potentially marginalizing large portions of the global population from accessing advanced AI services \citep{wang2024seaeval,zhu2024multilingual}.

This disparity has directly spurred research efforts to enhance multilingual capabilities of LLMs. Current approaches are predominantly based on data-driven post-training paradigm, such as multilingual instruction tuning \citep{zhang2023bayling,zhu2023extrapolating,ustun2024aya} or continual pre-training \citep{cui2023efficient,kuulmets2024teaching,jaavid2024romansetu}, which primarily seeks to elicit cross-lingual alignment \citep{schuster2019cross} or inject multilingual knowledge with the supervision signals from \emph{external} datasets.

While this paradigm is widely embraced and demonstrates certain successes, it faces several significant limitations. (1) The efficacy often depends on training with large-scale multilingual datasets \citep{cui2023efficient,zhu2023extrapolating}, which incur large computational overhead. (2) It overly emphasizes alignment across languages, neglecting the modeling of language-specific features, which exacerbates the off-target issue \citep{zhang2020improving}. As a result, the model often struggles to generate accurate responses in the intended language when prompted \citep{lai2024llms,sennrich2024mitigating}. (3) The model’s performance in languages it previously handled well is risking at catastrophic forgetting \citep{mccloskey1989catastrophic}.

In this work, we seek to provide a new perspective on addressing the aforementioned limitations by exploring and manipulating the \emph{internal} representation within the language-related latent spaces of LLMs \citep{zou2023representation,park2024linear}. Taking the enhancement of multilingual capabilities for English-centric LLMs as an example. This is based on the intuitive idea that the well-established English representations in existing English-centric LLMs can act as a pivot to improve the performance of other languages. More specifically, for the target language to be enhanced, its \emph{language-agnostic} semantic representations should be \emph{pulled close} to those of English, facilitating cross-lingual alignment. Meanwhile, the \emph{language-specific} linguistic representations should be \emph{pushed away} from English to preserve the unique representation of each language. Also, during this process, it is crucial to ensure that the English pivot representation remains unchanged to effectively prevent catastrophic forgetting.

\begin{figure*}
\centering
\resizebox{\linewidth}{!}{%
        \includegraphics{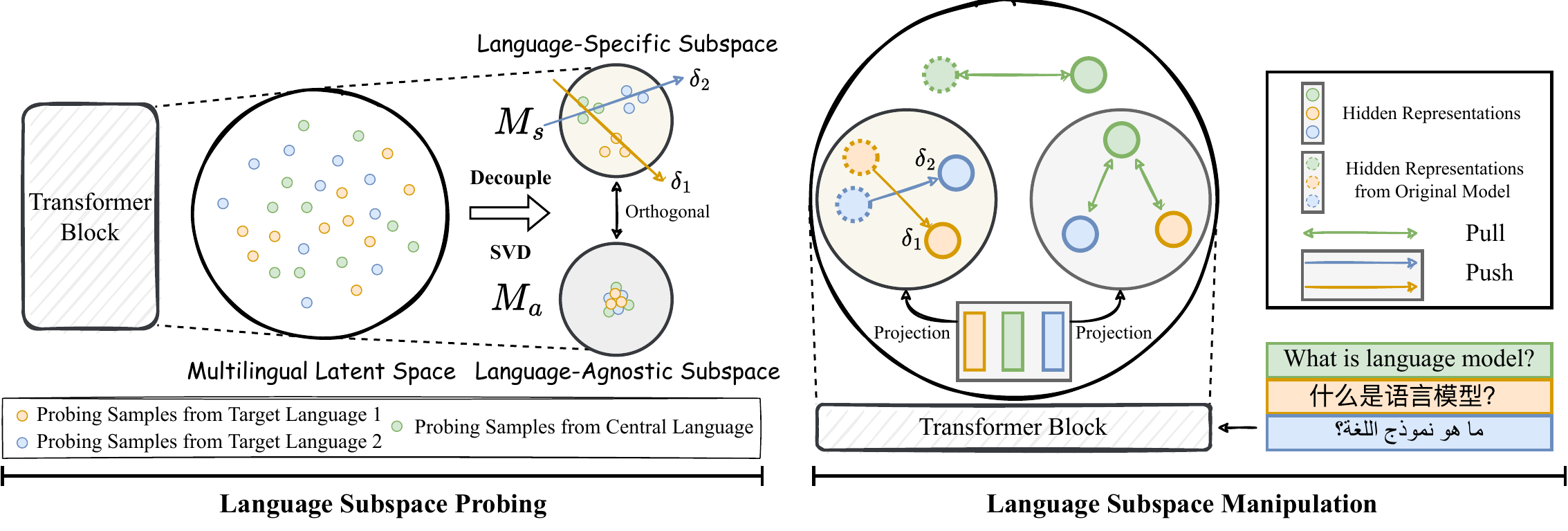}
    }
\caption{The overall architecture of our proposed \textsc{Lens} for multilingual enhancement. (1) In the LSP, we begin by decomposing the multilingual latent space, which is formed by the representations of probing samples from both the target and central languages. Using singular value decomposition (SVD), we separate this space into two orthogonal components: a language-agnostic subspace, $\boldsymbol{M}_{a}$, and a language-specific subspace, $\boldsymbol{M}_{s}$. (2) Then in LSM, the parallel multilingual representations of the target languages are pushed toward their respective linguistic expression directions within $\boldsymbol{M}_{s}$, while being pulled closer to the central language in $\boldsymbol{M}_{a}$. Additionally, the representations of the central language are carefully constrained to remain largely intact.}
\label{model}
\end{figure*}

To achieve this, we propose \textsc{Lens}, a novel multi\textbf{\underline{L}}ingual \textbf{\underline{E}}nhancement method based on the hidden represe\textbf{\underline{N}}tations within language \textbf{\underline{S}}pace of LLMs. To be more specific, \textsc{Lens} comprises two stages: Language Subspace Probing (LSP) and Language Subspace Manipulation (LSM). During LSP, the multilingual hidden space within a single layer of the backbone are decoupled into two orthogonal components, a \emph{language-agnostic} subspace and a \emph{language-specific} subspace, via singular value decomposition. Then in LSM, we align the parallel multilingual input representations of the target language and the central language in the language-agnostic subspace. Simultaneously, the projection components of the target language within the language-specific space are pushed away from those of the central language, guiding the target language toward its distinct linguistic expression and ensuring the target language is properly expressed thereby mitigating the off-target issue. Finally, we align the central language’s current representations with its original ones to preserve its proficiency during multilingual enhancement. It is crucial to note that, building on recent findings that language-related parameters are primarily concentrated in the top layers of LLMs \citep{wendler-etal-2024-llamas}, \textsc{Lens} only updates the higher layers of the backbone with just a few hundred data points, exhibiting high resource efficiency.

We conduct extensive experiments on bilingual and multilingual enhancement setups, targeting languages from diverse linguistic families and varying resource levels. Results on three English-centric LLMs (LLaMA-3-8B-Instruct, LLaMA-3.1-8B-Instruct and Phi-3.5-mini-Instruct) show that \textsc{Lens} outperform baselines and open-source multilingual-enhanced LLMs on both multilingual comprehension and generation tasks. Notably, \textsc{Lens} achieves these improvements without compromising the central language’s strong capabilities, while requiring significantly less computational overhead, highlighting its effectiveness, efficiency, and scalability.

The main contributions of this work are summarized as follows: (1) We provide a novel perspective for the multilingual enhancement of LLMs with their internal language representation space leveraged. (2) We propose \textsc{Lens}, an efficient and effective multilingual enhancement method that operates within the language representation space of large language models. (3) Extensive experiments demonstrate the effectiveness, efficiency, scalability of our method to obtain multilingual enhanced chat-style backbones without sacrificing original central language performance.

\section{Methodology}

\subsection{Overview of \textsc{Lens}}
We propose \textsc{Lens}, a novel method for effective and efficient multilingual enhancement of LLMs based on their internal language representation spaces. The overall diagram of \textsc{Lens} is displayed in Figure \ref{model}, consisting of two key stages: (1) Language Subspace Probing (LSP) and (2) Language Subspace Manipulation (LSM). The subsequent section offers a detailed introduction to them.

\subsection{Language Subspace Probing}
In this section, we first introduce our method to decouple and probe the language-agnostic and language-specific subspace within a single model layer in an unsupervised manner.

Assuming we aim to enhance the multilingual capabilities of a backbone model for $L$ languages, which include one central language and $L - 1$ target languages to be enhanced. In each layer of the backbone, we can obtain a mean representation for each language $l$:
\begin{equation}
\boldsymbol{m}_l = \frac{1}{n} \sum_{i=1}^n \boldsymbol{e}_l^i
\end{equation}
where $\boldsymbol{e}_l^i \in \mathbb{R}^{d}$ is the embedding of the last token for the $i$-th sample in language $l$, and $n$ is the total number of samples for each language. Concatenating $\boldsymbol{m}_l$ of $L$ languages column-by-column results in the mean embedding matrix $\boldsymbol{M} \in \mathbb{R}^{d \times L}$ specifying the multilingual latent space.

Following previous works \citep{pires-etal-2019-multilingual,libovicky2020language,yang2021simple}, we hypothesize that such multilingual latent space $M$ could be decomposed into two orthogonal components (1) a language-agnostic subspace $\boldsymbol{M}_{a}$ representing what is commonly shared across languages and (2) a language-specific one $\boldsymbol{M}_{s}$ specifying on which different languages express different linguistic signals. Following \citet{piratla2020efficient,xie2022discovering}, the objective can be formulated as:
\begin{equation}\label{eq:objective}
    \begin{aligned}
    \min_{\boldsymbol{M}_{a}, \boldsymbol{M}_{s}, \boldsymbol{\Gamma}}
    \quad& \left\|\boldsymbol{M}-\boldsymbol{M}_{a} \boldsymbol{\mathbbm{1}}^{\top}-\boldsymbol{M}_{s} \boldsymbol{\Gamma}^{\top}\right\|_{F}^{2}\\
    \textrm{s.t.} \quad& \text{Span}\left(\boldsymbol{M}_{a}\right) \perp \text{Span}\left(\boldsymbol{M}_{s}\right),
    \end{aligned}
\end{equation}
where $\boldsymbol{M}_{a} \in \mathbb{R}^{d \times 1}$, $\boldsymbol{M}_{s} \in \mathbb{R}^{d \times r}$ and $\boldsymbol{\Gamma} \in \mathbb{R}^{L \times r}$ is the coordinates of language-specific signals along the subspace's $r$ components. And a lower dimensionality for $\boldsymbol{M}_{a}$ is reasonable because the semantic consistency across different languages can be captured in a simpler form. Meanwhile, $\boldsymbol{M}_{s}$ requires a higher dimensionality to account for the distinct features of each language.

The optimal solution of Equation \ref{eq:objective} can be computed efficiently via Singular Value Decomposition (SVD), where Algorithm \ref{alg:ours} in Appendix \ref{app:prob} presents the detailed procedure.

After obtaining the language-specific subspace $\boldsymbol{M}_s$, we aim to identify a direction of language expression within this subspace, which points from the projection of mean representation from target language $\boldsymbol{m}_l$ to that from central language $\boldsymbol{m}_c$. Formally, the linguistic language expression direction $\boldsymbol{\delta}_l \in \mathbb{R}^{d}$ for each target language $l$ is calculated as:
\begin{gather}
    \boldsymbol{\delta}_l = \boldsymbol{M}_s^T \boldsymbol{M}_s (\boldsymbol{m}_l - \boldsymbol{m}_c) 
\end{gather}

\subsection{Language Subspace Manipulation}
To eliminate the heavy reliance on hard-to-access high-quality multilingual datasets, we leverage the well-trained hidden representations of the central language in LLMs as a pivot to derive supervision signals for multilingual enhancement within the model’s internal language space.

First, we propose to pull parallel multilingual representations closer within the shared language-agnostic subspace $\boldsymbol{M}_a$. This allows us to directly inherit the well-established general capabilities of the central language. Formally, this goal is accomplished by projecting multilingual representations (at the position of the last token) onto the subspace $\boldsymbol{M}_a$, with the optimization objective defined as:
\begin{gather}
    \mathcal{L}_1 = \left\|\boldsymbol{M}_a^T \boldsymbol{M}_a (\boldsymbol{x}_l - \boldsymbol{x}_c)\right\|^{2}
\end{gather}
where $\boldsymbol{x}_l$ and $\boldsymbol{x}_c$ are parallel multilingual representations from target language $l$ and central one. 

Second, to ensure that each target language can be accurately expressed and to alleviate the off-target issue, we need to push the multilingual representations in the language-specific subspace $\boldsymbol{M}_s$ towards their respective language-specific expression directions. This can be achieved through the projection onto the subspace $\boldsymbol{M}_s$ and optimizing the following objective:
\begin{gather}
    \mathcal{L}_2 = \left\|\boldsymbol{M}_s^T \boldsymbol{M}_s (\boldsymbol{x}_l - \boldsymbol{x}_l^{\text{ref}}) - \lambda_l \boldsymbol{\delta}_l \right\|^{2}
\end{gather}
where $\boldsymbol{x}_l^{\text{ref}}$ is the representation of target language $l$ obtained from original reference model and $\lambda_l$ is a scalar of push strength for the corresponding language. The above process can be interpreted as directing the language-specific representations of each target language to shift a specific distance from their original positions toward a direction that enables accurate expression.

Finally, to ensure that the capabilities of the central language are not compromised and maintain a stable alignment objective for the target language, we constrain the representations of central language to remain predominantly intact:
\begin{gather}
    \mathcal{L}_3 = \left\|\boldsymbol{x}_c - \boldsymbol{x}_c^{\text{ref}}\right\|^{2}
\end{gather}
where $\boldsymbol{x}_c^{\text{ref}}$ is the representation of central language $c$ obtained from original reference model.

The final optimization objective of \textsc{Lens} is:
\begin{gather}
    \mathcal{L} = \lambda_1 \mathcal{L}_1 + \mathcal{L}_2 + \lambda_3 \mathcal{L}_3
\end{gather}
where $\lambda_1$ and $\lambda_3$ are hyper-parameters to balance the impact of these two losses.

\section{Experiments}
\subsection{Experimental Setup}
\paragraph{Models} We select three English-centric LLMs: LLaMA-3-8B-Instruct \citep{llama3modelcard}, LLaMA-3.1-8B-Instruct \citep{dubey2024llama} and Phi-3.5-mini-instruct \citep{abdin2024phi}, to fully validate the efficacy and scaleability of our \textsc{Lens}. Notably, although the latter two models undergo more extensive multilingual alignment training, English continues to exhibit the strongest performance.

\paragraph{Languages to be Enhanced} We conduct experiments in both bilingual and multilingual settings to address various multilingual enhancement needs.

In the bilingual setting, English (En) serves as the central language, while Chinese (Zh) is the target language for expansion. Chinese is selected due to its growing prominence in the academic focus on multilingual enhancement for LLMs.

In the multilingual setting, we select six target languages, reflecting diverse linguistic families and resource levels. The high-resource languages are Chinese (Zh) and Japanese (Jp); the medium-resource languages are Korean (Ko) and Arabic (Ar); and the low-resource languages are Bengali (Bn) and Swahili (Sw), with English (En) continuing to serve as the central language.

It is important to note that these target languages are classified as \emph{out-of-scope} in the official model card of the above LLMs, which further underscores their relevance for enhancement.

\paragraph{Training Data} We sample 300 data points from Aya Dataset \citep{ustun2024aya} for each language to probe the language space and 200 data points from the Bactrian-X dataset \citep{li2023bactrian} per language to manipulate the language space. Please see Appendix \ref{app:data} for more details of these datasets.

\paragraph{Benchmarks} To comprehensively measure the efficacy of our \textsc{Lens} on various multilingual tasks, we employ 5 mainstream benchmarks for evaluation, which can be categorized into multilingual understanding and multilingual generation:

\textbf{Multilingual Understanding}: (1) \textbf{XCOPA} \citep{ponti2020xcopa}, (2) \textbf{XWinograd} \citep{muennighoff2023crosslingual}, (3) \textbf{XStoryCloze} \citep{lin2022few} and (4) \textbf{M-MMLU} \citep{hendrycks2021measuring,lai2023okapi}. \textbf{Accuracy} is the evaluation metric and we randomly sample up to 1,000 data points from each benchmark for evaluation.

\textbf{Multilingual Generation}: (5) \textbf{MT-Bench} \citep{zheng2023judging}: A benchmark for open-ended generation to evaluate a model’s ability to follow multi-turn instructions. The evaluation follows the \textbf{LLM-as-a-judge} approach, where GPT-4o is prompted to assign a score directly to a single response on a scale of 1 to 10. It is essential to highlight that the languages targeted for enhancement, as mentioned above, are all within the capability range of GPT-4o, especially given that its official model card \citep{openai2024gpt4o} emphasizes support for low-resource languages such as Swahili (Sw) and Bengali (Bn). This underscores the validity and reliability of the evaluation approach. We employ \textbf{Language Fidelity} \citep{holtermann2024evaluating} as a metric to assess the consistency between input and output languages, offering a clear measure of how effectively off-target issues are mitigated.

Please refer to Appendix \ref{app:benchmark} for the detailed description of the benchmarks.

\subsection{Baseline Methods}
For comparison, we consider the following baseline methods that enhance LLMs' multilingual capabilities using multilingual instruction fine-tuning technique: (1) \textbf{xSFT \& xSFT-Full} \citep{ouyang2022training}, (2) \textbf{QAlign} \citep{zhu-etal-2024-question}, (3) \textbf{SDRRL} \citep{zhang-etal-2024-enhancing-multilingual} and (4) \textbf{CLA} \citep{li2024improving}.

Please refer to Appendix \ref{app:baseline} for the detailed description of the baseline methods.

\begin{table*}[h]
\centering
\scriptsize
\setlength{\extrarowheight}{0pt}
\resizebox{\linewidth}{!}{
\begin{tabular}{l | c c c c c c c c c c | c c}
\toprule
\textbf{}        & \multicolumn{10}{c|}{\textbf{Multilingual Understanding}} & \multicolumn{2}{c}{\textbf{Multilingual Generation}} \\
\textbf{}        & \multicolumn{2}{c}{\textbf{XCOPA}} & \multicolumn{2}{c}{\textbf{XWinograd}} & \multicolumn{2}{c}{\textbf{XStoryCloze}} & \multicolumn{2}{c}{\textbf{M-MMLU}} & \multicolumn{2}{c|}{\textbf{AVG.}} & \multicolumn{2}{c}{\textbf{MT-Bench}} \\
\textbf{} & En & Zh & En & Zh & En & Zh & En & Zh & En & Zh & En & Zh \\
\midrule
LLaMA-3 & - & 83.40 & 63.50 & 54.37 & 95.40 & 88.90 & 64.90 & 49.40 & 74.60 & 69.02 & 6.99 (100\%) & 2.72 (43.75\%) \\
\midrule
xSFT & - & \cellcolor{mycolor_green}\underline{87.20} & \cellcolor{mycolor_green}\textbf{64.30} & \cellcolor{mycolor_green}\underline{63.49} & \cellcolor{mycolor_green}\textbf{95.10} & \cellcolor{mycolor_green}\underline{90.60} & \cellcolor{mycolor_red}\underline{62.80} & \cellcolor{mycolor_red}46.10 & \cellcolor{mycolor_red}\underline{74.07} &\cellcolor{mycolor_green}\underline{71.85} & \cellcolor{mycolor_red}4.79 (100\%) & \cellcolor{mycolor_green}2.94 (88.75\%) \\
xSFT-Full & - & \cellcolor{mycolor_green}84.60 & \cellcolor{mycolor_red}58.80 & \cellcolor{mycolor_green}60.11 & \cellcolor{mycolor_red}93.50 & \cellcolor{mycolor_green}90.30 & \cellcolor{mycolor_red}60.60 & \cellcolor{mycolor_red}43.20 & \cellcolor{mycolor_red}70.97 & \cellcolor{mycolor_green}69.55 & \cellcolor{mycolor_red}5.80 (100\%) & \cellcolor{mycolor_green}\underline{4.44} (\underline{92.50\%}) \\
QAlign & - & \cellcolor{mycolor_red}52.20 & \cellcolor{mycolor_red}55.10 & \cellcolor{mycolor_red}47.02 & \cellcolor{mycolor_red}89.20 & \cellcolor{mycolor_red}71.90 & \cellcolor{mycolor_red}56.40 & \cellcolor{mycolor_red}34.00 & \cellcolor{mycolor_red}66.90 & \cellcolor{mycolor_red}51.28 & \cellcolor{mycolor_red}3.59 (100\%) & \cellcolor{mycolor_red}1.23 (37.50\%) \\
SDRRL & - &\cellcolor{mycolor_red}85.20 & \cellcolor{mycolor_green}64.80 & \cellcolor{mycolor_green}55.95 & \cellcolor{mycolor_red}92.60 & \cellcolor{mycolor_red}84.30 & \cellcolor{mycolor_red}63.80 & \cellcolor{mycolor_red}{47.80} & \cellcolor{mycolor_red}73.73 & \cellcolor{mycolor_red}68.31 & \cellcolor{mycolor_red}\underline{6.60} (100\%) & \cellcolor{mycolor_green}3.84 (73.75\%) \\
CLA &- &\cellcolor{mycolor_green}85.60 &\cellcolor{mycolor_red}61.70 &\cellcolor{mycolor_green}56.70 &\cellcolor{mycolor_green}\underline{95.00} &\cellcolor{mycolor_green}89.80 &\cellcolor{mycolor_green}\textbf{64.70} & \cellcolor{mycolor_red}\textbf{48.90} &\cellcolor{mycolor_red}73.80 &\cellcolor{mycolor_green}70.26 &\cellcolor{mycolor_red}6.47 (100\%) &\cellcolor{mycolor_green}{4.41} (81.25\%)\\
\midrule
\textsc{Lens} (Ours) & - & \cellcolor{mycolor_green}\textbf{87.60} & \cellcolor{mycolor_green}\underline{63.80} & \cellcolor{mycolor_green}\textbf{66.67} & \cellcolor{mycolor_red}{94.70} & \cellcolor{mycolor_green}\textbf{91.80} & \cellcolor{mycolor_green}\underline{64.40} & \cellcolor{mycolor_red}\underline{48.60} & \cellcolor{mycolor_green}\textbf{74.30} & \cellcolor{mycolor_green}\textbf{73.67} & \cellcolor{mycolor_green}\textbf{7.21} (100\%) & \cellcolor{mycolor_green}\textbf{5.77} (\textbf{97.50\%}) \\
\midrule
\midrule
LLaMA-3.1 & - & 90.40 & 64.10 & 68.65 & 95.80 & 91.40 & 69.30 & 52.50 & 76.40 & 75.74 & 7.31 (100\%) & 5.38 (93.75\%) \\
\midrule
xSFT & - & \cellcolor{mycolor_red}\underline{88.00} & \cellcolor{mycolor_green}{63.70} & \cellcolor{mycolor_red}\underline{67.46} & \cellcolor{mycolor_green}\underline{96.20} & \cellcolor{mycolor_green}\textbf{92.70} & \cellcolor{mycolor_red}{68.10} & \cellcolor{mycolor_green}\textbf{53.10} & \cellcolor{mycolor_red}{76.00} & \cellcolor{mycolor_red}{75.32} &\cellcolor{mycolor_red}5.33 (100\%) & \cellcolor{mycolor_red}3.32 (90.00\%) \\
xSFT-Full & - & \cellcolor{mycolor_red}86.80 & \cellcolor{mycolor_red}60.40 & \cellcolor{mycolor_red}62.50 & \cellcolor{mycolor_red}90.60 & \cellcolor{mycolor_red}83.80 & \cellcolor{mycolor_red}66.10 & \cellcolor{mycolor_red}49.90 & \cellcolor{mycolor_red}72.37 & \cellcolor{mycolor_red}70.75 & \cellcolor{mycolor_red}\underline{6.02} (100\%) & \cellcolor{mycolor_red}{4.18} (\underline{92.50\%}) \\
QAlign & - & \cellcolor{mycolor_red}55.00 & \cellcolor{mycolor_red}56.00 & \cellcolor{mycolor_red}48.02 & \cellcolor{mycolor_red}94.10 & \cellcolor{mycolor_red}52.30 & \cellcolor{mycolor_red}64.10 & \cellcolor{mycolor_red}33.50 & \cellcolor{mycolor_red}71.40 & \cellcolor{mycolor_red}47.20 & \cellcolor{mycolor_red}4.13 (100\%) & \cellcolor{mycolor_red}2.65 (83.75\%) \\
SDRRL &- & \cellcolor{mycolor_red}87.20 & \cellcolor{mycolor_red}63.20 & \cellcolor{mycolor_red}58.83 & \cellcolor{mycolor_green}95.30 & \cellcolor{mycolor_red}89.80 & \cellcolor{mycolor_red}63.50 & \cellcolor{mycolor_red}45.30 & \cellcolor{mycolor_red}74.00 & \cellcolor{mycolor_red}70.31 & \cellcolor{mycolor_red}6.49 (100\%) & \cellcolor{mycolor_red}3.14 (58.75\%) \\
CLA &- &\cellcolor{mycolor_red}89.00 &\cellcolor{mycolor_green}63.30 &\cellcolor{mycolor_red}65.90 &\cellcolor{mycolor_green}\textbf{94.20} &\cellcolor{mycolor_green}\underline{90.50} &\cellcolor{mycolor_green}\textbf{67.40} &\cellcolor{mycolor_red}52.50 & \cellcolor{mycolor_green}\textbf{77.20} &\cellcolor{mycolor_red}\underline{75.36} &\cellcolor{mycolor_red}6.49 (100\%) &\cellcolor{mycolor_red}\underline{4.49} (88.75 \%)\\
\midrule
\textsc{Lens} (Ours) & - & \cellcolor{mycolor_green}\textbf{90.20} & \cellcolor{mycolor_green}\underline{64.60} & \cellcolor{mycolor_green}\textbf{69.44} & \cellcolor{mycolor_green}{95.90} & \cellcolor{mycolor_green}{91.80} & \cellcolor{mycolor_green}\underline{69.10} & \cellcolor{mycolor_green}\underline{52.60} & \cellcolor{mycolor_green}\underline{76.53} & \cellcolor{mycolor_green}\textbf{76.01} & \cellcolor{mycolor_green}\textbf{7.41} (100\%) & \cellcolor{mycolor_green}\textbf{5.96} (\textbf{93.75\%}) \\
\bottomrule
\end{tabular}
}
\caption{Detailed results on the multilingual understanding and multilingual generation benchmarks under the bilingual setting (English and Chinese). The values in parentheses represent language fidelity. Results highlighted \mycolorbox{mycolor_green}{in green} indicate an improvement or performance of central language comparable (within a gap of 0.5) to the original backbone, while those highlighted \mycolorbox{mycolor_red}{in red} signal a decline in performance relative to the original backbone. The best and second-best results in our method and baselines are in bold and underlined, respectively.}
\label{tab:llama3_bilingual_all_results}
\end{table*}

\subsection{Implementation Details}
Our experiments are implemented with PyTorch \citep{paszke2019pytorch} and Transformer library \citep{wolf2020transformers} on a single NVIDIA A800-SXM4-80GB GPU. The training duration is set to one epoch with the learning rate of 1e-5 and batch size of 8 across all backbones. For more detailed settings, please refer to the Appendix \ref{app:implement}.

\subsection{Overall Results}
\label{subsec:main_experiment}

Table \ref{tab:llama3_bilingual_all_results} present the performance comparison between \textsc{Lens} and recent multilingual enhancement baseline methods under bilingual settings. For additional results under multilingual configuration across all three backbones, please see Appendix \ref{app:results_mul}. From the results across all backbones, we have drawn the following key insights:

\textbf{\textsc{Lens} achieves comprehensive improvements across diverse languages with different resource levels.} It outperforms baselines on both multilingual understanding and generation benchmarks, effectively mitigating the off-target issue. In contrast, baselines primarily favor multilingual understanding, offering minimal or no improvement in generation tasks. In some cases, they compromise language fidelity, exacerbating the off-target problem. Finally, \textsc{Lens} safeguards the central language from catastrophic forgetting. We also compare \textsc{Lens} with LoRA-based SFT, as recent work suggests that it can effectively prevent catastrophic forgetting \citep{huang2024mindmerger}. Detailed results and discussions are provided in Appendix \ref{app:results_lora}.

\textbf{Using the central language representations within the backbone as a supervision signal proves more effective, efficient and scalable.} The key distinction between \textsc{Lens} and baseline methods lies in how multilingual performance is enhanced: \textsc{Lens} relies on the model’s internal representation of the central language, while baseline methods depend on external data. This difference make baselines not only fail to improve the target languages but also lead to performance degradation. However, the Aya Dataset and Bactrain-X datasets we used are already considered high-quality multilingual resources, widely employed and proven effective in boosting multilingual capabilities in previous models such as mT5 and LLaMA-2 \citep{li2023bactrian,ustun2024aya}. This highlights that for current extensively trained LLMs such as LLaMA-3 (which has been trained on over 15T data), an over-reliance on external supervision signals may fall short of scalability needs \citep{cao2024towards}. In addition, the results in Table \ref{tab:cost} of Appendix \ref{app:main_exp} demonstrate that \textsc{Lens} achieves the best performance with minimal computational overhead.

\begin{figure*}
\centering
\resizebox{\linewidth}{!}{%
            \includegraphics{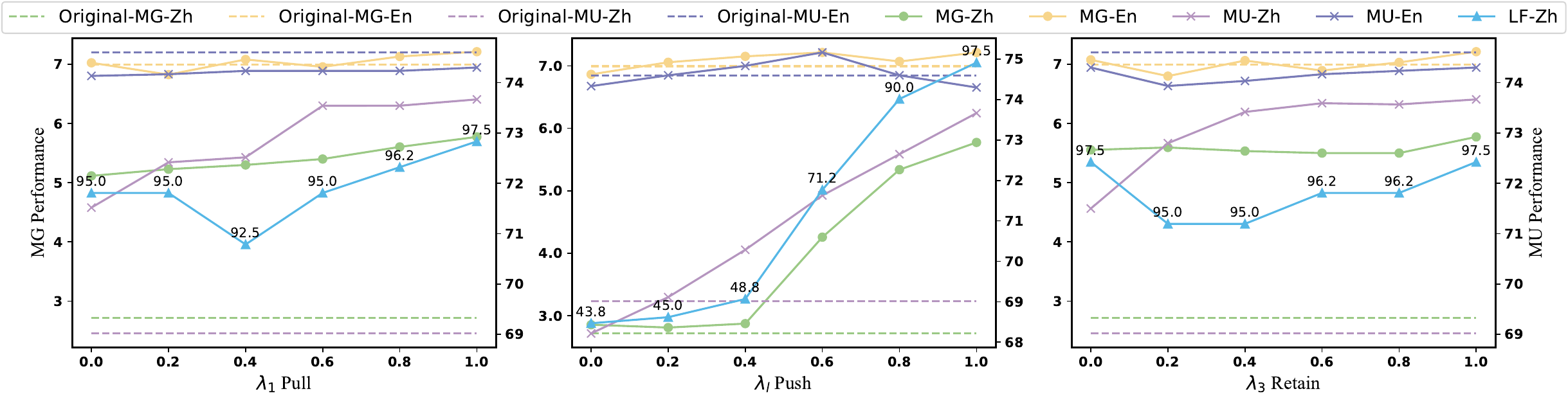}
    }
\caption{The ablation results to verify the effectiveness and impact of different optimization objectives in LSM. MU Performance stands for the average performance on all multilingual understanding benchmarks, while MG Performance is the results on MT-Bench. LF represents language fidelity.}
\label{fig:ablation}
\end{figure*}

\begin{table}[t!]
\centering
\setlength{\extrarowheight}{0pt}
\resizebox{\linewidth}{!}{
  \begin{tabular}{ccccccc}
    \toprule
    & \textbf{Zh} & \textbf{Jp} & \textbf{Ar} & \textbf{Ko} & \textbf{Bn} & \textbf{Sw}\\
    \midrule
    \rowcolor{gray!20} \multicolumn{7}{c}{\textit{COMET scores on X-to-English Tasks}} \\
    LLaMA-3 &85.40	&86.15	&84.77	&86.07	&85.51	&78.15 \\
    \midrule
    xSFT	&70.41	&72.40	&67.09	&72.43	&59.52	&73.56 \\
    QAlign	&85.52	&85.26	&83.11	&84.96	&83.13	&73.66 \\
    SDRRL	&44.78	&45.73	&40.87	&45.29	&45.05	&41.51 \\
    CLA &85.28	&85.35	&85.11	&85.15	&84.36	&77.84 \\
    \midrule
    \textsc{Lens} &\textbf{85.64}	&\textbf{86.23}	&\textbf{85.15}	&\textbf{86.07}	&\textbf{85.67}	&\textbf{80.05} \\
    \midrule
    \midrule
    \rowcolor{gray!20} \multicolumn{7}{c}{\textit{COMET scores on English-to-X Tasks}} \\
    LLaMA-3	&85.28	&88.32	&76.51	&84.53	&80.14	&71.44 \\
    \midrule
    xSFT	&83.78	&82.22	&74.30	&81.08	&73.40	&58.48 \\
    QAlign	&61.65	&58.66	&49.41	&57.16	&41.10	&50.96 \\
    SDRRL	&62.52	&57.65	&43.83	&64.11	&68.74	&60.00 \\
    CLA &80.19	&85.75	&72.50	&82.45	&56.73	&55.36 \\
    \midrule
    \textsc{Lens} &\textbf{85.59}	&\textbf{88.47}	&\textbf{79.52}	&\textbf{85.77}	&\textbf{80.20}	&\textbf{71.88} \\
    \bottomrule
  \end{tabular}
  }
  \caption{Results of baseline methods and our \textsc{Lens} on FLORES-101 benchmark.}
  \label{tab:flores}
\end{table}

\section{Analysis}
\subsection{Ablation Study}
\label{subsec:ablation}
We conduct ablation studies to validate the effectiveness of the three optimization objectives in LSM. The results under the bilingual enhancement setting with LLaMA-3-8B-Instruct are shown in Figure \ref{fig:ablation}, leading to the following key findings:

\textbf{Aligning multiple languages in language-agnostic subspaces primarily enhances multilingual comprehension rather than generation.} As the coefficient $\lambda_1$ of the alignment loss $\mathcal{L}_1$ increases, Chinese comprehension improves, while its generation ability remains slightly unaffected.

\textbf{Separating language representations in language-specific subspaces is crucial for multilingual performance.} As shown in the middle of Figure \ref{fig:ablation}, increasing $\lambda_l$ significantly boosts both comprehension and generation in Chinese, yielding greater benefits than mere language alignment. This suggests that the conventional approach of aligning languages \citep{cao2020multilingual,zhu2023extrapolating,li2024improving,hua2024mothello} may be insufficient for fully optimizing multilingual LLMs. Further discussion in Appendix \ref{app:dis_sah} elaborates on this insight.

To further illustrate the benefit of the distinction between representations of different languages within the language-specific subspace, we conduct additional experiments on the multilingual machine translation task and results are shown in Table \ref{tab:flores}. The consistent performance improvements achieved by \textsc{Lens} confirm that it captures the subtle linguistic nuances required for precise and fluent translations. Detailed experimental setups and analysis are provided in Appendix \ref{app:mul_trans}.

\textbf{Maintaining stable representations of the central language provides reliable alignment supervision.} As shown in the rightmost part of Figure \ref{fig:ablation}, removing the retention objective for English leads to a significant drop in Chinese performance, likely due to misalignment caused by changes in English representations. However, since our modifications only affect the upper layers (layers 31 and 32) of LLaMA-3-8B-Instruct, most parameters remain frozen, preventing catastrophic forgetting of English capabilities. Further analysis in \S\ref{subsec:layer} confirms that increasing the number of updated layers does not negatively impact English performance, highlighting the efficacy of the retention objective.

\begin{figure*}
    \centering
    \subfigure[The impact of manipulating different backbone layers on multilingual performance enhancement.]{\includegraphics[width=0.32\textwidth]
    {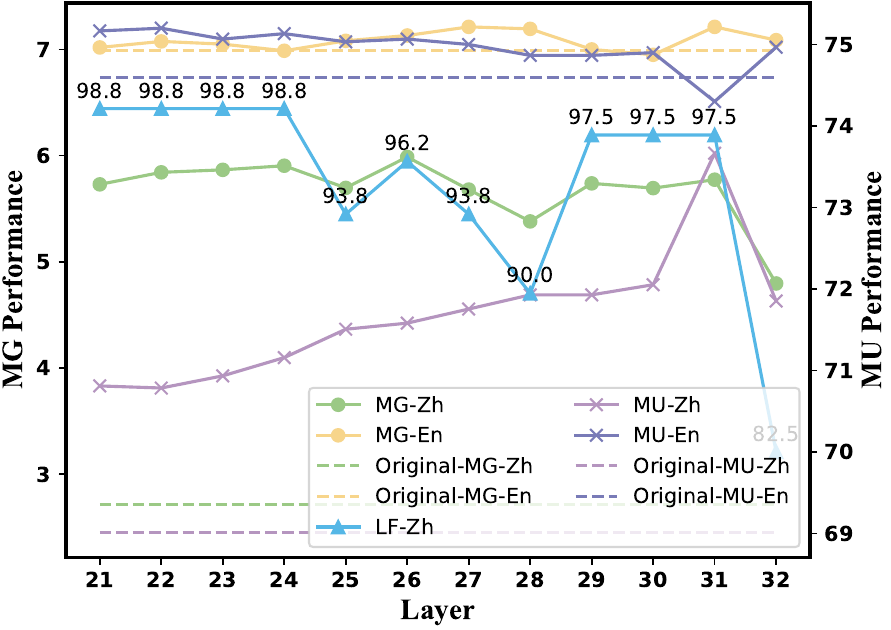}
    \label{subfig:layer_analysis}}
    \subfigure[The impact of varying the number of training data per language.]{\includegraphics[width=0.32\textwidth]
    {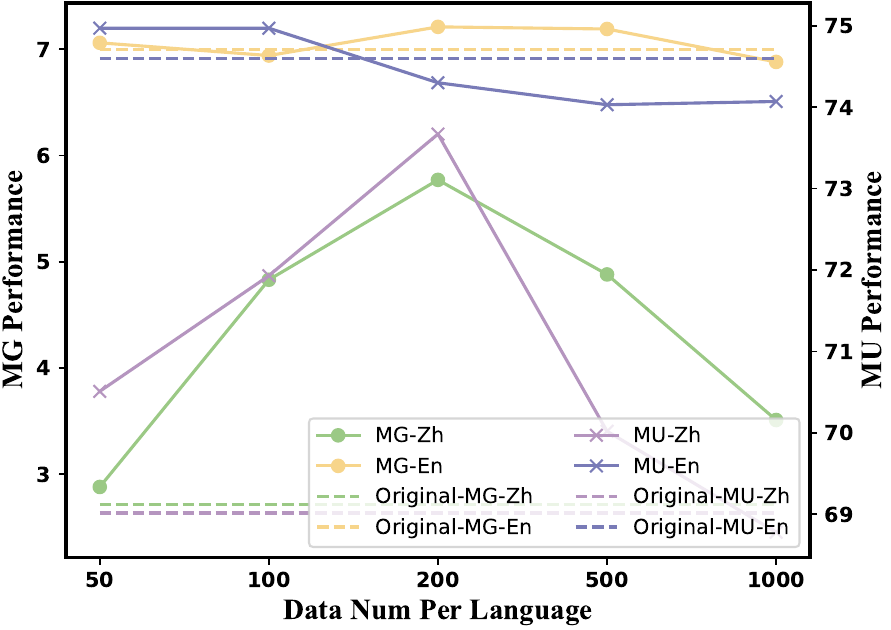}
    \label{subfig:data_num}}
    \subfigure[Comparison between the bilingual enhanced model via our \textsc{Lens} with Chinese-LLaMA series.]{\includegraphics[width=0.31\textwidth]
    {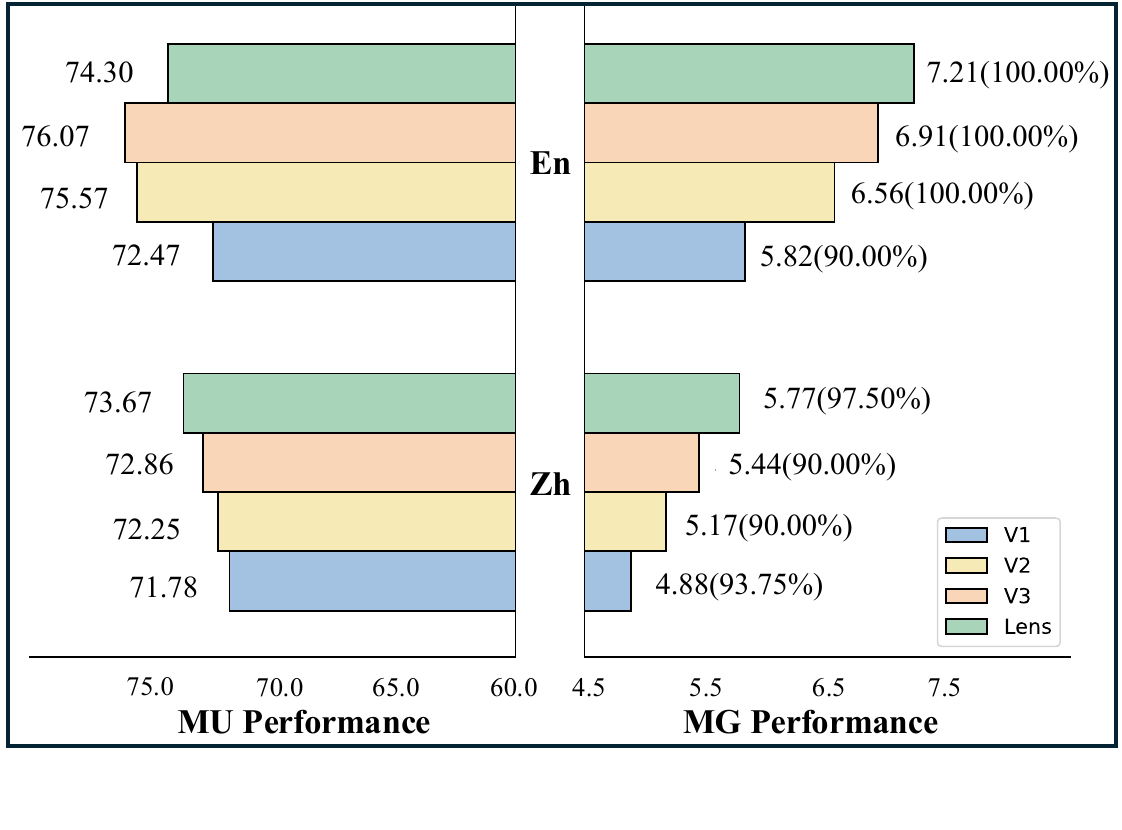}
    \label{subfig:comprare_chinese}}
    \caption{(a) The impact of varying the number of manipulated layers. (b) The impact of training data volume. \\ (c) Comparison with open-source multilingual-enhanced LLMs.}
    \label{fig:visual_all}
\end{figure*}

\begin{figure}
\centering
\resizebox{\linewidth}{!}{%
            \includegraphics{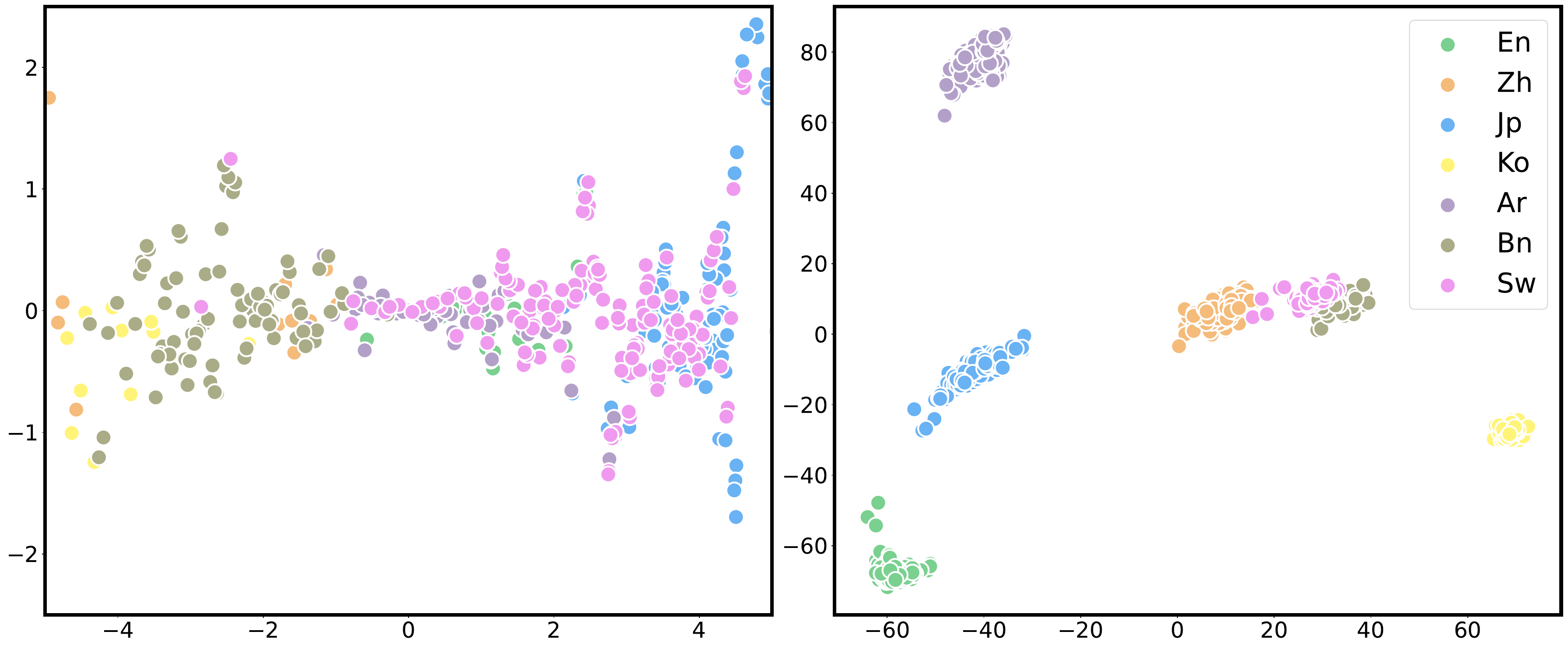}
    }
\caption{The PCA visualization of multilingual representations projected in the obtained language-agnostic subspace (right) and the language-specific (left) subspace. The backbone model is LLaMA-3-8B-Instruct after multilingual enhanced with \textsc{Lens}.}
\label{fig:visual}
\end{figure}

\subsection{Impact of Varying the Number of Manipulated Layers and Training Data}
\label{subsec:layer}
Building on recent studies on LLM interpretability \citep{zhao2024large,zhong2024beyond}, we focus on updating only the upper layers of the backbone. Figure \ref{subfig:layer_analysis} examines the impact of modifying different layers under the bilingual enhancement setting with LLaMA-3-8B-Instruct. The horizontal axis denotes the starting layer for updates, with the final layer as the default endpoint.

\textbf{Intermediate-layer processing in English benefits multilingual understanding.} If the target language is partitioned into the language-specific subspace too early, it weakens multilingual comprehension. However, modifying only the final layer is insufficient for enhancing either understanding or generation, as language-specific information remains underdeveloped, leading to off-target errors.

\textsc{Lens} further validates the conclusions of existing works on LLM interpretability and applies these findings to multilingual enhancement. Please refer to Appendix \ref{app:layer} for more analysis.

Results in Figure \ref{subfig:data_num} indicate that increasing the amount of training data leads to diminishing returns for \textsc{Lens}. This finding reinforces our claim that for extensively pre-trained LLMs such as LLaMA-3 (trained on over 15T tokens), over-reliance on more training data falls short of meeting scalability needs. Instead of focusing on larger training datasets, it is more critical to identify supervision signals that are both reliable and scalable. This directly motivates us to seek internal supervision from the central language with the backbone itself.

\subsection{Comparison with Open-Source Multilingual-Enhanced LLMs}
In \S\ref{subsec:main_experiment}, we compare with \emph{reproducible} baselines. Additionally, we extend our comparisons to open-source LLMs that leverage private datasets and large-scale post-training to improve multilingual performance. In particular, we focus on the Chinese-LLaMA-3 series, which builds on LLaMA-3 series to enhance Chinese capabilities and includes three versions. For a detailed description of these models, please refer to Appendix \ref{app:open_models}.

Results and the resource consumption of different methods are presented in Figure \ref{subfig:comprare_chinese} and Table \ref{tab:cost} in Appendix \ref{app:main_exp}, respectively. The resulting model applied with \textsc{Lens} is identical to the one utilized for bilingual enhancement in Table \ref{tab:llama3_bilingual_all_results}. \textsc{Lens} demonstrates more comprehensive enhancement of the Chinese capabilities with extremely low resource overhead compared to these three models. Also, all the data leveraged by \textsc{Lens} is publicly accessible, which eliminates the need for laboriously gathering extensive high-quality multilingual datasets and makes it easily shareable with the community.

\subsection{Visualization Analysis}
To further confirm whether \textsc{Lens} manipulates language representations within different language subspaces as anticipated, as shown in Figure \ref{fig:visual}, we perform Principal Component Analysis (PCA) to visualize the projection of multilingual representations in the language-agnostic and -specific subspace. Parallel inputs in seven languages are sourced from the MultiQ datasets \citep{holtermann2024evaluating}. Results indicate that representations of different languages converge within a narrow range in the language-agnostic subspace, while forming distinct clusters in the language-specific subspace, supporting our claim. This also highlights the advantages of \textsc{Lens} in delivering transparent, controllable, and interpretable solutions for the multilingual enhancements of LLMs.

\subsection{Impact of Language Proximity}
In \S\ref{subsec:main_experiment}, we focus on languages that are out-of-scope and under-represented in the backbone, selected based on their typological and script differences from the central language, English. To further highlight the generalizability of \textsc{Lens}, we expand our analysis to include three additional languages: Spanish, French, and German. These languages, being typologically closer to English, are already better represented in the backbone model. As illustrated in Table \ref{tab:llama3_es_fr_de} of Appendix \ref{app:proxi}, \textsc{Lens} still exhibit notable improvements. For a more comprehensive discussion, please refer to Appendix \ref{app:proxi}.

\section{Related Works}
\paragraph{Multilingual Large Language Model} Pretraining on multilingual data is a common approach to gain multilingual capabilities \citep{conneau2019cross,xue2020mt5,lin2022few,shliazhko2022mgpt,wei2023polylm,xue2022byt5,le2023bloom,blevins2024breaking}. However, due to the uneven distribution of data in pretraining corpora, current LLMs or MLLMs exhibit uneven language capabilities, with most models heavily biased towards English \citep{jiang2023mistral,llama3modelcard,abdin2024phi}. Moreover, pretraining from scratch is computationally intensive. These limitations have directly sparked research into expanding or enhancing the language capabilities of current LLMs or MLLMs.

\paragraph{Multilingual Enhancement for LLMs} Current methods for multilingual enhancement of LLMs can be categorized into two types: 1) prompt-based methods and 2) post-training-based methods.

The former focuses on leveraging the LLMs' own translation capabilities to translate low-resource language inputs into the central language, and then generating a response \citep{shi2023language,huang2023not,qin2023cross,etxaniz2024multilingual,zhang-etal-2024-autocap}. However, \citet{liu2024translation} reveal the limitations of these methods, showing they are not optimal for real-world scenarios and highlighting the necessity of more comprehensive multilingual enhancement.

The latter aims to conduct further multilingual post-training to inject language knowledge or achieve cross-lingual alignment, including ways of continual pre-training \citep{zhang2021cpm,cui2023efficient,chen2023lifelong,lin2024mala,kuulmets2024teaching,jaavid2024romansetu} and instruction tuning \citep{muennighoff2023crosslingual,chen2023phoenix,indurthi2024improving,ahuja2024sphinx,lai-nissim-2024-mcot,zhang-etal-2024-enhancing-multilingual,zhu-etal-2024-question,li2024improving,zhao2024adamergex}. For example, \citet{cui2023efficient} attempt to inject Chinese knowledge into LLaMA by conducting continual pre-training on a large-scale Chinese corpus, while \citet{zhu2023extrapolating} focus more on building language alignment through cross-lingual instruction tuning and translation training.

Our proposed \textsc{Lens} stands out from existing methods in that we seek multilingual supervision signals from the \emph{internal} language representation space of the LLMs, rather than relying heavily on \emph{external} multilingual datasets as in the above methods, which offers fresh insights and new opportunities for enhancing the multilingual capabilities of LLMs both efficiently and effectively. For more discussions on the theoretical foundation of language-agnostic and language-specific subspaces within LLMs, please refer to Appendix \ref{app:theory}.

\paragraph{Representation Engineering} Editing or manipulating representation within LLMs has garnered increasing attention due to its transparency and lightweight properties \citep{zou2023representation}. This is theoritically rooted from Linear Representation Hypothesis \citep{mikolov2013linguistic,nanda2023emergent,park2024linear}. Building upon this, existing works attempt to edit representations at inference time to develop models that are more truthful \citep{li2023inference,campbell2023localizing,zhang2024truthx}, and harmless \citep{lee2024mechanistic,uppaal2024detox}. We expand and implement this paradigm for the multilingual enhancement of LLMs by focusing on representations during the training phase, ensuring that the efficiency of LLMs remains unaffected during the inference phase.

\section{Conclusion}

In this paper, we introduce \textsc{Lens}, a novel method designed for the effective, efficient and comprehensive multilingual enhancement of large language models (LLMs). \textsc{Lens} first decouples the multilingual hidden spaces of the backbone into two orthogonal components: a language-agnostic subspace and a language-specific subspace. Then taking well-established representations of the central language as a pivot, representations of target languages are pulled closer and pushed away from them in language-agnostic subspace and language-specific subspace, respectively.  Experimental results on 3 representative cutting-edge LLMs demonstrate that \textsc{Lens} outperforms baseline methods with much lower training costs, underscoring its efficacy, efficiency and scalability.

\section*{Limitation and Future Work}
Despite our \textsc{Lens} achieving comprehensive and efficient multilingual enhancement, there are still limitations and future directions worth exploring.

First, due to limited computational resources, our experiments are not conducted on larger-scale models (larger than 8B). This remains a valuable direction to apply \textsc{Lens} on larger LLMs.

Second, our current operations on language representation are still relatively coarse-grained. Future work could delve into more specific parameter areas for finer operations.

Finally, as we find that relying too much on external datasets to enhance multilingual capabilities may be limited, we instead seek higher quality supervision signals from within the model itself. Future work could consider combining these two paradigms by incorporating data selection strategies \citep{albalak2024survey,liu2024makes}, thereby providing higher quality multilingual supervision signals to the model from both internal and external sources.

\section*{Ethical Considerations}
This work is conducted solely for academic research purposes, aiming to enhance the multilingual capabilities of large language models in a responsible and interpretable manner. Our methods focus on improving cross-lingual alignment and representation learning without introducing biases or harmful content. We do not intend for this research to be used in applications that could lead to misinformation, discrimination, or unethical manipulation of language models. Additionally, we adhere to ethical AI principles, ensuring that our approach respects linguistic diversity and maintains fairness across different languages.

\bibliography{custom}

\newpage

\appendix

\section{Probing for Language Subspace}
\label{app:prob}

The optimal solution of Equation~\ref{eq:objective} can be computed efficiently via Singular Value Decomposition (SVD).
Algorithm~\ref{alg:ours} presents the detailed procedure.
Readers interested in more details can consult the proof provided in \citet{xie2022discovering}.
The only hyperparameter $r < L$ controls the amount of language-specific information captured by the identified subspace.
The larger $r$ is, the more language-specific signals we can identify.

\input{algo/ours}

\section{Training Data}
\label{app:data}
The multilingual data used for the language subspace probing stage is sourced from the Aya Dataset \citep{ustun2024aya}, a human-annotated, non-parallel multilingual instruction fine-tuning dataset with 204,000 instances in 65 languages. For the language subspace manipulation stage, we rely on parallel multilingual data from the Bactrian-X dataset \citep{li2023bactrian}, which contains 3.4 million instruction-response pairs in 52 languages. These pairs are generated by translating 67,000 English instructions (derived from alpaca-52k \citep{alpaca} and dolly-15k) into 51 languages using the Google Translate API, and then obtaining natural responses from ChatGPT.

\section{Multilingual Benchmarks}
\label{app:benchmark}
We comprehensively measure the efficacy of our \textsc{Lens} on various multilingual tasks, including 5 mainstream benchmarks for evaluation. They can be categorized into the evaluation of multilingual understanding and multilingual generation.

For multilingual understanding:
\begin{itemize}
    \item \textbf{XCOPA} \citep{ponti2020xcopa}:\footnote{\url{https://huggingface.co/datasets/cambridgeltl/xcopa}} A benchmark to evaluate the ability of machine learning models to transfer commonsense reasoning across languages. The dataset is the translation and re-annotation of the English COPA \citep{roemmele2011choice} and covers 11 languages from 11 families and several areas around the globe. The dataset is challenging as it requires both the command of world knowledge and the ability to generalise to new languages. In our experimental setup, this benchmark covers both Chinese (Zh) and Swahili (Sw).
    \item \textbf{XWinograd} \citep{muennighoff2023crosslingual}:\footnote{\url{https://huggingface.co/datasets/Muennighoff/xwinograd}} A well-established tool for evaluating coreference resolution (CoR) and commonsense reasoning (CSR) capabilities of computational models. The dataset is the translation of the English Winograd Schema datasets and it adds 488 Chinese schemas from CLUEWSC2020 \citep{xu2020clue}, totaling 6 languages. Formulated as a fill-in-a-blank task with binary options, the goal is to choose the right option for a given sentence which requires commonsense reasoning. In our experimental setup, this benchmark covers English (En), Chinese (Zh) and Japanese (Jp).
    \item \textbf{XStoryCloze} \citep{lin2022few}:\footnote{\url{https://huggingface.co/datasets/juletxara/xstory_cloze}} A commonsense reasoning framework for evaluating story understanding, story generation, and script learning. The dataset consists of the professionally translated version of the English StoryCloze dataset (Spring 2016 version) to 10 non-English languages. The dataset is challenging and is designed to evaluate story understanding, story generation, and script learning. In our experimental setup, this benchmark covers English (En), Chinese (Zh), Arabic (Ar) and Swahili (Sw).
    \item \textbf{M-MMLU} \citep{hendrycks2021measuring,lai2023okapi}:\footnote{\url{https://huggingface.co/datasets/alexandrainst/m_mmlu}} A benchmark designed to measure knowledge acquired during pretraining by evaluating models exclusively in zero-shot and few-shot settings. The datasets is a machine translated version of the MMLU dataset by GPT-3.5-turbo and covers 34 languages. This is a massive multitask test consisting of multiple-choice questions from various branches of knowledge. To attain high accuracy on this test, models must possess extensive world knowledge and problem solving ability. In our experimental setup, this benchmark covers English (En), Chinese (Zh), Arabic (Ar), Korean (Ko), and Swahili (Sw).
\end{itemize}

For multilingual generation:
\begin{itemize}
    \item \textbf{MT-Bench} \citep{zheng2023judging}: The dataset is designed for open-ended generation to evaluate a model’s ability to follow multi-turn instructions. In our experimental setup, this benchmark covers English (En), Chinese (Zh), Arabic (Ar), Japanese (Jp), Korean (Ko), Swahili (Sw) and Bengali (Bn). We collect data in English\footnote{\url{https://huggingface.co/datasets/HuggingFaceH4/mt_bench_prompts}}, Japanese\footnote{\url{https://huggingface.co/datasets/shi3z/MTbenchJapanese}}, Korean\footnote{\url{https://huggingface.co/datasets/StudentLLM/Korean_MT-Bench_questions}}, and Arabic\footnote{\url{https://huggingface.co/spaces/QCRI/mt-bench-ar/tree/main/data/mt_bench_ar}} from huggingface, and Chinese\footnote{\url{https://github.com/HIT-SCIR/huozi}} from github. In addition, we use GPT-4o to translate the English data into Swahili and Bengali, and performed manual proofreading to ensure correctness.
\end{itemize}

\section{Baseline Methods}
\subsection{Multilingual Enhancement Baselines}
\label{app:baseline}
For comparison, we consider the following baseline methods that enhance LLMs' multilingual capabilities using multilingual instruction fine-tuning technique: (1) \textbf{xSFT \& xSFT-Full} \citep{ouyang2022training}: xSFT performs multilingual instruction fine-tuning using the same data volume as our \textsc{Lens}. In contrast, xSFT-Full utilizes the full dataset for each target language from the Aya Collection and Bactrian-X. (2) \textbf{QAlign} \citep{zhu-etal-2024-question}: It explores the benefits of question alignment, where the model is trained to translate inputs into English by finetuning on X-English parallel question data. (3) \textbf{SDRRL} \citep{zhang-etal-2024-enhancing-multilingual}: It is based on self-distillation from resource-rich languages that effectively improve multilingual performance by leveraging self-distillated data. (4) \textbf{CLA} \citep{li2024improving}: It aligns the internal sentence representations across different languages via multilingual contrastive learning and aligns outputs by following cross-lingual instructions in the target language.

\subsection{Open-Source Multilingual-Enhanced LLMs}
\label{app:open_models}

The Chinese-LLaMA-3 series, which builds on LLaMA-3 series to enhance Chinese capabilities and includes three different versions:

\begin{itemize}
    \item \textbf{Chinese-LLaMA-3-Instruct-V1}:\footnote{\url{https://huggingface.co/hfl/llama-3-chinese-8b-instruct}} This model is continually pre-trained on 120GB of Chinese text and fine-tuned with 500 million carefully curated instruction data points, based on the LLaMA-3-8B. These training datasets is not available to the public.
    \item \textbf{Chinese-LLaMA-3-Instruct-V2}:\footnote{\url{https://huggingface.co/hfl/llama-3-chinese-8b-instruct-v2}} This version is directly fine-tuned on the same 500 million instruction data points using the LLaMA-3-8B-Instruct model.
    \item \textbf{Chinese-LLaMA-3-Instruct-V3}:\footnote{\url{https://huggingface.co/hfl/llama-3-chinese-8b-instruct-v3}} This model is created by merging V1, V2, and the original LLaMA-3-8B-Instruct, followed by fine-tuning on 5,000 instruction data points.
\end{itemize}

\begin{table}
\centering
\small
\begin{tabular}{l | c c }
\toprule
& Manipulated Layer & $\lambda_{\text{Zh}}$ \\
\midrule
LLaMA-3-8B-Instruct&31 &1 \\
LLaMA-3.1-8B-Instruct&30 &0.05 \\
Phi-3.5-mini-Instruct&27 &0.25 \\
\bottomrule
\end{tabular}
\caption{Detailed hyper-parameter settings for bilingual enhancement. The number under the column of Manipulated Layer represents the starting point of the layers where manipulation is applied, with the default endpoint being the final layer.}
\label{tab:hyper_bilingual}
\end{table}

\begin{table*}
\centering
\small
\begin{tabular}{l | c c c c c c c }
\toprule
& Manipulated Layer & $\lambda_{\text{Zh}}$ &$\lambda_{\text{Jp}}$ &$\lambda_{\text{Ko}}$ &$\lambda_{\text{Ar}}$ &$\lambda_{\text{Bn}}$ &$\lambda_{\text{Sw}}$ \\
\midrule
LLaMA-3-8B-Instruct&29 &1 &0.6 &1 &0.5 &0.2 &0.2 \\
LLaMA-3.1-8B-Instruct&30 &0.01 &0.01 &0.03 &0.01 &0.01 &0.01 \\
Phi-3.5-mini-Instruct&29 &0.2 &0.2 &0.2 &0.2 &0.15 &0.15 \\
\bottomrule
\end{tabular}
\caption{Detailed hyper-parameter settings for multilingual enhancement. The number under the column of Manipulated Layer represents the starting point of the layers where manipulation is applied, with the default endpoint being the final layer.}
\label{tab:hyper_multilingual}
\end{table*}

\section{Implementation Details}
\label{app:implement}
Our experiments are implemented with PyTorch \citep{paszke2019pytorch} and Transformer library \citep{wolf2020transformers} on a single NVIDIA A800-SXM4-80GB GPU. The training duration is set to one epoch with the learning rate of 1e-5, cosine learning rate scheduler with warm up ratio of 0.05 and batch size of 8 across all backbones.  And all backbones are trained with their official chat template with $\lambda_1 = 1$ and $\lambda_3 = 1$. The hyper-parameter $r$ specifying the dimension of language-specific subspace in language subspace probing stage is set to $L-1$, where $L$ is the total number of languages participated in this process. We use GlotLID \citep{kargaran2023glotlid} to identify the response language to obtain the language fidelity. GlotLID is an open-source language identification model that supports more than 1,600 languages. GlotLID returns iso\_636\_9 language codes, which we manually map to the language codes in this work.

More detailed hyper-parameter settings for bilingual and multilingual enhancement across different backbones are listed in Table \ref{tab:hyper_bilingual} and Table \ref{tab:hyper_multilingual}.

Further, we carefully evaluate the official implementations of all baselines, in order to make the comparison as fair as possible. All baselines are trained using the same language pairs as those in \textsc{Lens}. For SDRRL and CLA, we adhere to their respective configurations to reconstruct the training data from these language pairs. We strictly follow the hyper-parameter settings in their original code. If this could not reach the expected performance, we carry out the hyper-parameter search of the learning rate and batch size.

\begin{table*}
\centering
\scriptsize
\setlength{\extrarowheight}{0pt}
\resizebox{\linewidth}{!}{
\begin{tabular}{l | c c c c c c c c c c | c c}
\toprule
\textbf{}        & \multicolumn{10}{c|}{\textbf{Multilingual Understanding}} & \multicolumn{2}{c}{\textbf{Multilingual Generation}} \\
\textbf{}        & \multicolumn{2}{c}{\textbf{XCOPA}} & \multicolumn{2}{c}{\textbf{XWinograd}} & \multicolumn{2}{c}{\textbf{XStoryCloze}} & \multicolumn{2}{c}{\textbf{M-MMLU}} & \multicolumn{2}{c|}{\textbf{AVG.}} & \multicolumn{2}{c}{\textbf{MT-Bench}} \\
\textbf{} & En & Zh & En & Zh & En & Zh & En & Zh & En & Zh & En & Zh \\
\midrule
Phi-3.5 & - & 81.40 & 75.80 & 67.70 & 95.40 & 89.40 & 71.70 & 47.30 & 81.00 & 71.40 & 6.18 (100\%) & 4.92 (90.50\%) \\
\midrule
xSFT & - & \cellcolor{mycolor_red}\underline{80.80} & \cellcolor{mycolor_green}\textbf{77.20} & \cellcolor{mycolor_green}\textbf{69.64} & \cellcolor{mycolor_green}95.40 & \cellcolor{mycolor_green}89.40 & \cellcolor{mycolor_green}{71.70} & \cellcolor{mycolor_red}46.80 & \cellcolor{mycolor_green}\textbf{81.43} &\cellcolor{mycolor_green}\underline{71.66} & \cellcolor{mycolor_red}5.29 (100\%) & \cellcolor{mycolor_red}3.31 (88.75\%) \\
xSFT-Full & - & \cellcolor{mycolor_red}80.40 & \cellcolor{mycolor_red}73.10 & \cellcolor{mycolor_red}65.67 & \cellcolor{mycolor_red}95.20 & \cellcolor{mycolor_red}88.20 & \cellcolor{mycolor_green}\textbf{71.90} & \cellcolor{mycolor_red}44.70 & \cellcolor{mycolor_red}80.07 & \cellcolor{mycolor_red}69.74 & \cellcolor{mycolor_red}5.25 (100\%) & \cellcolor{mycolor_red}\underline{3.84} (87.50\%) \\
QAlign & - & \cellcolor{mycolor_red}78.00 & \cellcolor{mycolor_red}69.60 & \cellcolor{mycolor_red}58.73 & \cellcolor{mycolor_red}95.10 & \cellcolor{mycolor_red}84.70 & \cellcolor{mycolor_red}70.80 & \cellcolor{mycolor_red}46.60 & \cellcolor{mycolor_red}78.50 & \cellcolor{mycolor_red}67.01 & \cellcolor{mycolor_red}5.28 (100\%) & \cellcolor{mycolor_red}3.15 (88.75\%) \\
SDRRL & - & \cellcolor{mycolor_green}\underline{81.80} & \cellcolor{mycolor_green}\underline{76.30} & \cellcolor{mycolor_red}66.87 & \cellcolor{mycolor_green}\underline{95.60} & \cellcolor{mycolor_green}\underline{90.20} & \cellcolor{mycolor_red}71.60 & \cellcolor{mycolor_red}\underline{46.90} & \cellcolor{mycolor_green}\underline{81.17} & \cellcolor{mycolor_green}71.44 & \cellcolor{mycolor_red}\underline{6.15} (100\%) & \cellcolor{mycolor_red}4.03 (90.00\%) \\
CLA &- &\cellcolor{mycolor_red}80.40 &\cellcolor{mycolor_green}76.50 &\cellcolor{mycolor_red}66.50 &\cellcolor{mycolor_green}95.70 &\cellcolor{mycolor_green}89.40 &\cellcolor{mycolor_green}71.70 &\cellcolor{mycolor_red}47.10 &\cellcolor{mycolor_green}81.30 &\cellcolor{mycolor_red}70.85 &\cellcolor{mycolor_red}6.08 (100\%)&\cellcolor{mycolor_red}4.26 (90.00\%) \\
\midrule
\textsc{Lens} (Ours) & - & \cellcolor{mycolor_green}\textbf{82.60} & \cellcolor{mycolor_green}75.80 & \cellcolor{mycolor_green}\underline{68.73} & \cellcolor{mycolor_green}\textbf{95.60} & \cellcolor{mycolor_green}\textbf{90.60} & \cellcolor{mycolor_green}\underline{71.80} & \cellcolor{mycolor_green}\textbf{47.40} & \cellcolor{mycolor_green}81.07 & \cellcolor{mycolor_green}\textbf{72.33} & \cellcolor{mycolor_green}\textbf{6.44} (100\%) & \cellcolor{mycolor_green}\textbf{5.16} (\textbf{92.50\%}) \\
\bottomrule
\end{tabular}
}
\caption{Detailed results on the multilingual understanding and multilingual generation benchmarks with Phi-3.5-mini-Instruct backbone under the bilingual setting (English and Chinese). Accuracy serves as the evaluation metric for multilingual understanding, while GPT-4o ratings (on a scale of 1 to 10) are provided for MT-Bench. The values in parentheses represent language fidelity. Results highlighted \mycolorbox{mycolor_green}{in green} indicate an improvement or performance comparable to the original backbone, while those highlighted \mycolorbox{mycolor_red}{in red} signal a decline in performance relative to the original backbone. The best and second-best results in our method and baselines are in bold and underlined.} 
\label{tab:phi3.5_bilingual_all_results}
\end{table*}

\begin{table*}
\centering
\scriptsize
\begin{tabular}{l | c c c c c c c c c }
\toprule
&Lens	&xSFT	&xSFT-Full	&SDRRL	&QAlign &CLA	&V1	&V2	&V3\\
\midrule
Training time &2m08s	&5m33s	&192m35s	&11m30s	&12m03s &12m33s &- &- &-			 \\
Trainable parameters rate &5.43\%	&100.00\%	&100.00\% &100.00\% &100.00\% &100.00\% &13.08\% &13.08\%	&- \\
Instruction data &1K	&1K	&111.5K	&4K	&1K &1K	&5M	&5M	&5K\\
Pre-training data &- &- &- &- &- & - &120G &- &- \\
\bottomrule
\end{tabular}
\caption{Resource consumption of different multilingual enhancement methods under the bilingual enhancement setup. The backbone model is LLaMA-3-8B-Instruct.}
\label{tab:cost}
\end{table*}

\begin{figure*}
\centering
\resizebox{\linewidth}{!}{%
            \includegraphics{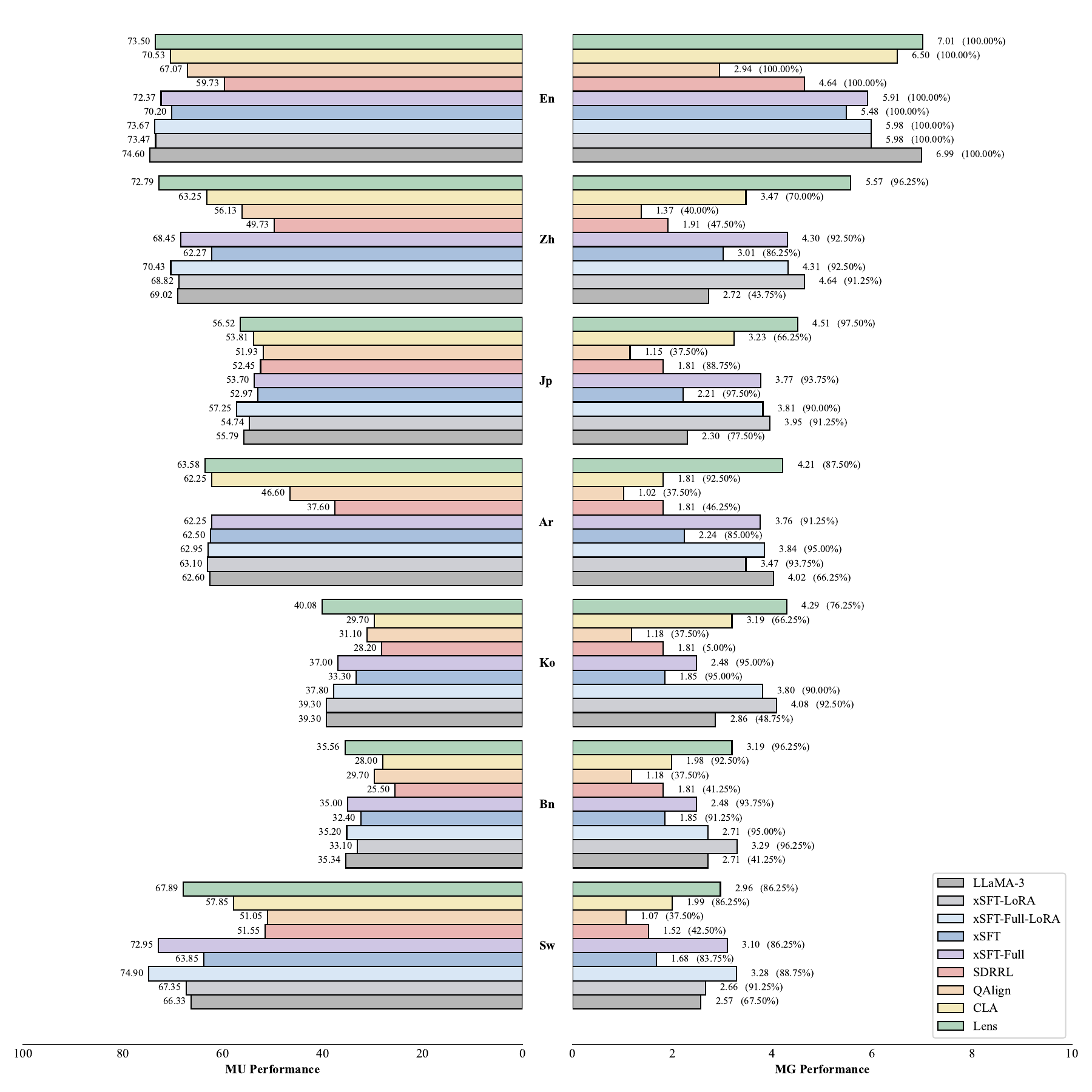}
    }
\caption{Results on the multilingual understanding and generation benchmarks with LLaMA-3-8B-Instruct backbone under the multilingual setting.}
\label{fig:multilingual_all_llama3}
\end{figure*}

\begin{figure*}
\centering
\resizebox{\linewidth}{!}{%
            \includegraphics{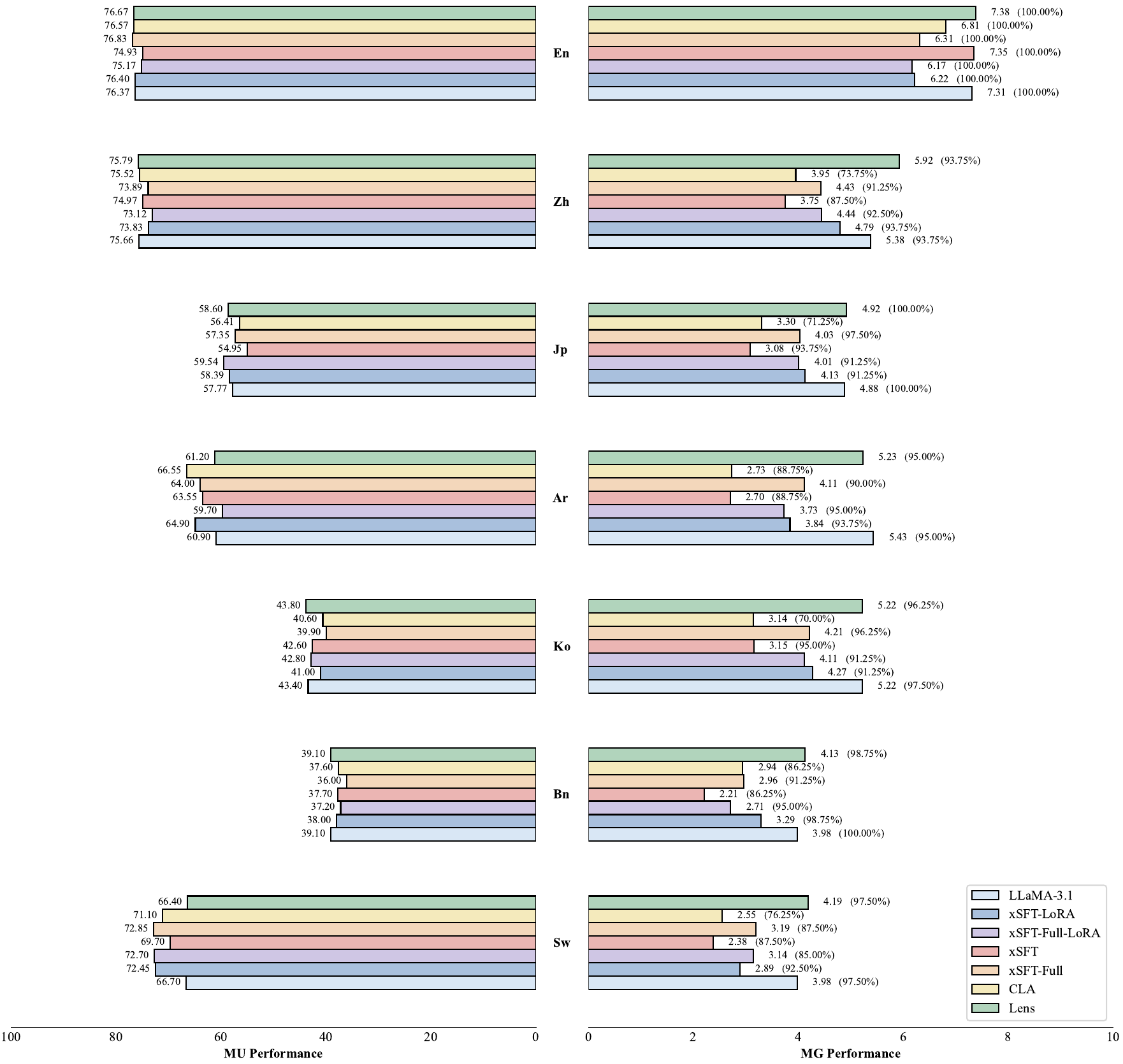}
    }
\caption{Results on the multilingual understanding and generation benchmarks with LLaMA-3.1-8B-Instruct backbone under the multilingual setting.}
\label{fig:multilingual_all_llama3.1}
\end{figure*}

\begin{figure*}
\centering
\resizebox{\linewidth}{!}{%
            \includegraphics{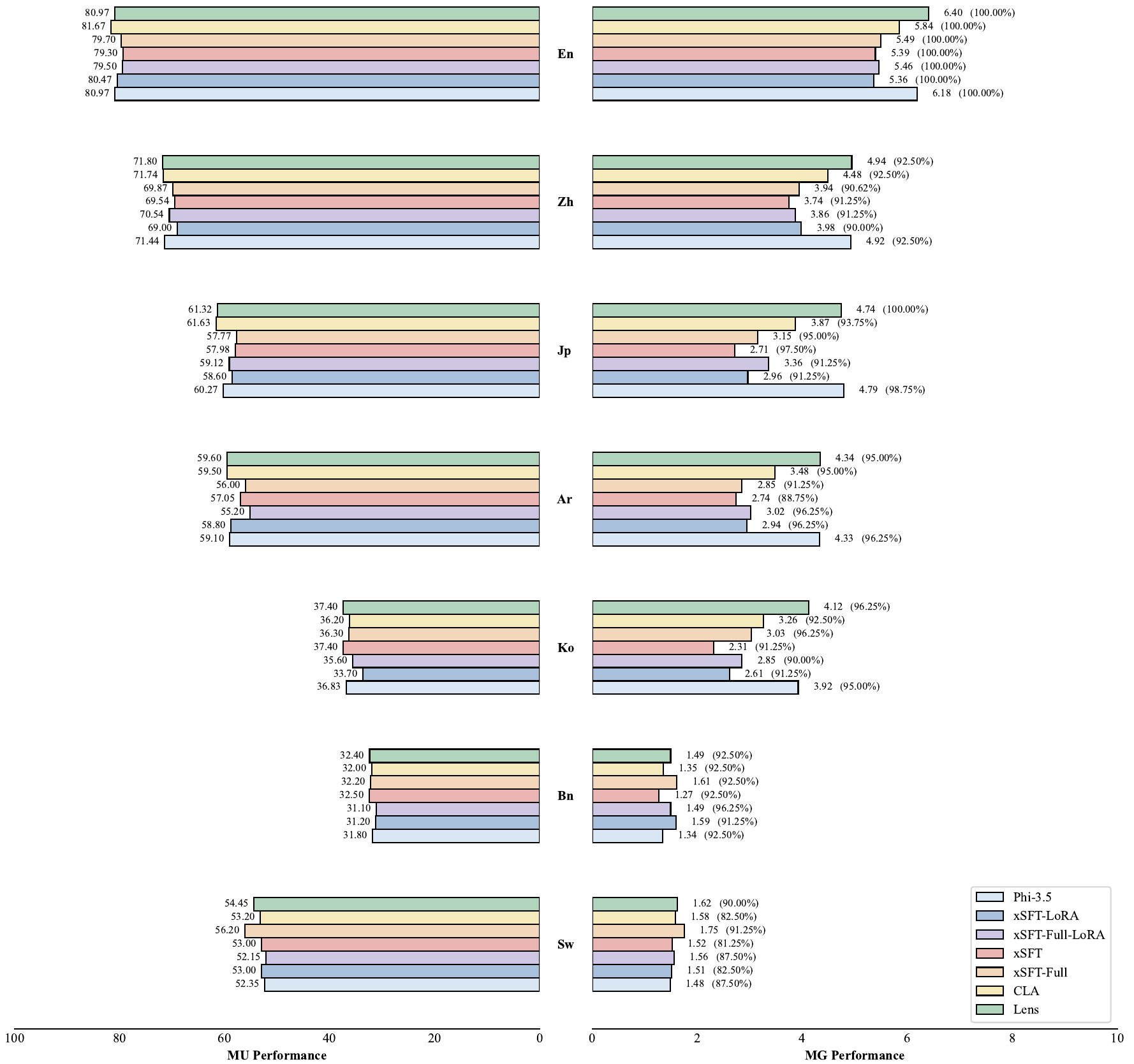}
    }
\caption{Results on the multilingual understanding and generation benchmarks with Phi-3.5-mini-Instruct backbone under the multilingual setting.}
\label{fig:multilingual_all_phi3.5}
\end{figure*}

\section{Additional Experimental Results}
\label{app:main_exp}

\subsection{Results under Multilingual Settings}
\label{app:results_mul}

Results of the comparison between \textsc{Lens} and baseline methods on Phi-3.5-mini-Instruct under bilingual and multilingual setups are shown in Table \ref{tab:phi3.5_bilingual_all_results} and Figure \ref{fig:multilingual_all_phi3.5}, respectively. 
We report the results of multilingual enhancement settings on LLaMA-3-8B-Instruct and LLaMA-3.1-8B-Instruct are displayed in Figure \ref{fig:multilingual_all_llama3} and \ref{fig:multilingual_all_llama3.1}, respectively.

The results demonstrate that our \textsc{Lens} is still capable of achieving the comprehensive multilingual enhancement. Similarly, \textsc{Lens} continues to improve the model's multilingual generation capability, enhancing the quality of the model's responses in specific languages. However, the improvement in language fidelity is more pronounced in the LLaMA-3-8B-Instruct backbone than in the multilingual backbones (LLaMA-3.1-8B-Instruct and Phi-3.5-mini-Instruct), which the latter ones undergo more extensive multilingual alignment training. Notably, while the baseline method considerably decreases the language fidelity of the multilingual backbones, \textsc{Lens} has minimal impact on it. These extensive experimental results demonstrate that \textsc{Lens} can serve as an effective, efficient, and scalable multilingual enhancement solution. We hope that our method can provide inspiration for future work to seek multilingual supervision more from the LLM itself rather than heavily relying on external dataset.

We also note that xSFT-Full exhibits uneven performance gains, especially in Swahili (Sw).  We believe this may be attributed to the uneven quality of the training data. Specifically, the Bactrian-X dataset used for training derives its input from Google Translate and its output from GPT-3.5-turbo, meaning the dataset quality depends heavily on these two sources. As a result, inconsistencies in translation and generation quality can introduce noise, leading to uneven performance gains from data-driven post-training approaches like xSFT-Full. This highlights one of the key limitations of the current data-driven paradigms.

In contrast, \textsc{Lens} seeks supervision signals internally from the backbone itself, bypassing the need for extensive reliance on potentially noisy external datasets. This intrinsic approach allows \textsc{Lens} to achieve consistent improvements over the backbone model across a wide range of languages, demonstrating better scalability and robustness. We have also demonstrated this phenomenon in our experiments, showcasing \textsc{Lens}’s broader applicability. In our future work, we propose combining the \textsc{Lens} training paradigm with advancements in data selection and filtering methods. We believe this hybrid approach holds great potential for further enhancing multilingual performance.

\subsection{Comparison with LoRA-Based SFT}
\label{app:results_lora}

In Figures \ref{fig:multilingual_all_llama3}, \ref{fig:multilingual_all_llama3.1} and \ref{fig:multilingual_all_phi3.5}, We also perform comparison with LoRA-based \citep{hu2022lora} SFT as recent work suggests that LoRA-based SFT can effectively prevent catastrophic forgetting \citep{huang2024mindmerger}. Based on our experimental results, we derived the following key conclusions:

For preserving the central language’s capabilities, incorporating LoRA-based SFT is indeed more effective at preventing catastrophic forgetting than its full-parameter counterpart. However, it primarily protects multilingual understanding (MU) tasks while multilingual generation (MG) capabilities are also significantly affected.

For target language enhancement, LoRA-based SFT methods also show a trend for improving MU tasks more over MG tasks.

By contrast, our proposed \textsc{Lens} achieves a more comprehensive performance, simultaneously enhancing understanding and generation for target languages while maintaining both the understanding and generation capabilities of the central language across different base models.

\subsection{Broader Insights and Connection to Superficial Alignment Hypothesis}
\label{app:dis_sah}
Our experimental findings suggest that eliciting language-specific representations benefits multilingual capability more than aligning language-agnostic representations, which may also lend support to the superficial alignment hypothesis \citep{zhou2023lima,lin2024unlocking,yan2024exploring}. It posits that LLMs acquire their core knowledge and abilities during pretraining, while post-alignment training primarily guides the model towards a desirable subdistribution of formats to use when prompted. In the multilingual settings, this is specifically manifested in:

(1) Despite the imbalance distribution of training data in pretraining resources for different languages, the majority of language-agnostic knowledge is already well-comprehended and aligned during pretraining, especially for current LLMs exposed to super-large-scale pretraining corpora (e.g., over 15T tokens for LLaMA-3).

(2) Current post-alignment training, which disproportionately focuses on English data, limits other languages to a subdistribution aligned with English-specific formats.

Thus, further aligning multilingual representations may have less impact compared to stimulating language-specific expressiveness in the target languages, but both mechanisms contribute to performance improvement in our method, with separation playing a more significant role.

\subsection{Results on Multilingual Machine Translation}
\label{app:mul_trans}

We carry out evaluations on the FLORES-101 dataset \citep{goyal2022flores}. Specifically, we assess the bidirectional translation performance between the target language and English, reporting scores using the COMET metric with the WMT22-comet-da model \citep{rei2022comet}. The choice of target language and training settings align with the experimental setup used for multilingual enhancement on LLaMA-3-8B-Instruct.

The experimental results in Table \ref{tab:flores} demonstrate that \textsc{Lens} still effectively enhances the multilingual machine translation performance, further validating its robustness across diverse multilingual tasks. Additionally, methods like QAlign, SDRRL and CLA, which heavily rely on translation-based training for multilingual alignment, fall short in acquiring deep linguistic understanding for each language. Thus, their ignorance of language-specific modeling hinders their efficacy in multilingual machine translation that require fine-grained linguistic nuances necessary for accurate translations.

\subsection{Impact of Varying the Number of Manipulated Layers}
\label{app:layer}

Recent studies on the interpretability of LLMs has sought to reveal the mechanisms underlying their multilingual capabilities \citep{zhao2024large,zhong2024beyond}. A growing consensus suggests that language-specific parameters or neurons are primarily concentrated in the upper layers of these models, while the middle layers tend to process inputs from various languages using a shared and language-agnostic mechanism \citep{chen2023bigger,wendler-etal-2024-llamas,tang-etal-2024-language,kojima2024multilingual,zhang-etal-2024-unveiling-linguistic}. Drawing inspiration from this, our main experiments focus on performing updates solely within the upper layers of the backbone, resulting in a notable improvement in multilingual performance. In Figure \ref{subfig:layer_analysis}, we explore the effect of increasing the number of layers involved on the model’s multilingual enhancement. The horizontal axis represents the starting point of the layers where manipulation is applied, with the default endpoint being the final layer. This experiment is performed under the bilingual enhancement with LLaMA-3-8B-Instruct.

``Thinking'' in English at the intermediate layers is more favorable for improving multilingual understanding. If we partition representations of target language into the language-specific subspace too early at the middle layers, it may impair its multilingual understanding capability. On the contrary, inheriting more from the shared representations at the middle layers, while emphasizing language-specific representations only at the higher layers (where most language-specific parameters and neurons are concentrated), is more beneficial for enhancing multilingual performance.

It is important to note that modifying only the final layer does not significantly improve either multilingual understanding or generation. This is because language-specific information is not sufficiently enhanced, causing the model to suffer from off-target issues and struggle to represent specific languages accurately. The lack of improvement in multilingual understanding aligns with the findings in \S\ref{subsec:ablation}, which highlight the critical role of supervision provided by the Push loss ($\mathcal{L}_2$).

\textsc{Lens} further validates the conclusions of existing works on LLM interpretability and applies these findings to multilingual enhancement.

\begin{table}
\centering
\resizebox{\linewidth}{!}{
\begin{tabular}{c | c c c | c c c}
\toprule
\textbf{}        & \multicolumn{3}{c|}{\textbf{MU}} & \multicolumn{3}{c}{\textbf{MG}} \\
&Es &Fr &De &Es &Fr &De \\
\midrule
LLaMA-3 &53.50	&52.50	&56.50 &5.88	&5.27	&4.56 \\
\midrule
xSFT &50.80	&49.90	&56.00 &4.36	&4.47	&4.00 \\
xSFT-Full &50.90	&48.90	&51.50 &4.60	&4.42	&4.33 \\
SDRRL &48.90	&47.80	&50.30 &2.74	&3.06	&2.54 \\
QAlign &48.50	&46.60	&51.30 &2.88	&2.91	&2.31 \\
\midrule
\textsc{Lens} &\textbf{53.70}	&52.10	&\textbf{57.10} &\textbf{5.90}	&\textbf{5.63}	&\textbf{4.90} \\
\bottomrule
\end{tabular}
}
\caption{Detailed results on the multilingual understanding and multilingual generation benchmarks with LLaMA-3-8B-Instruct backbone for Spanish (Es), French (Fr), and German (De). Accuracy serves as the evaluation metric for multilingual understanding, while GPT-4o ratings (on a scale of 1 to 10) are provided for MT-Bench. The values in parentheses represent language fidelity. The best and second-best results in our method and baselines are in bold and underlined.} 
\label{tab:llama3_es_fr_de}
\end{table}

\subsection{Impact of Language Proximity}
\label{app:proxi}

Here are our supplemented experimental results based on LLaMA-3-8B-Instruct to further highlight the generalizability of \textsc{Lens}, where these 3 languages, Spanish (Es), French (Fr), and German (De), are typologically and scripturally closer to English. For Multilingual Understanding (MU) evaluation, we adopt M-MMLU dataset which covers all 4 languages En, Es, Fr and De. And Multilingual Generation (MG) evaluation is performed on MT-Bench. Results are shown in Table \ref{tab:llama3_es_fr_de}.

The horizontal comparison in Table \ref{tab:llama3_es_fr_de} and Figure \ref{fig:multilingual_all_llama3} reveals that it demonstrates effectiveness across languages from different typological families, further highlighting its generalizability and adaptability to diverse linguistic characteristics.

\section{Additional Related Works}
\label{app:theory}

Here we provide additional discussion on the theoretical foundation of language-agnostic and language-specific subspaces, dividing it into two aspects: linguistic theory and LLM interpretability.

\paragraph{Linguistic Theory} From a linguistic standpoint, the idea of separating representations into language-agnostic and language-specific spaces is grounded in established theories of language universals and typology. Language-agnostic features align with universal linguistic structures, such as shared syntactic patterns or semantic primitives \citep{greenberg1963universals,comrie1989language}, while language-specific features capture unique aspects like phonology, morphology, or syntax \citep{croft2002typology,cotterell2016morphological}. These distinctions have also been studied in computational linguistics, such as in multilingual embeddings \citep{artetxe2018robust} and cross-lingual representation learning \citep{ruder2019survey}, supporting the conceptual basis in \textsc{Lens}.

\paragraph{LLM Interpretability} Recent interpretability studies have provided compelling evidence that LLMs internally encode language-agnostic and language-specific subspaces. For example, specific neurons or groups of neurons have been identified as responsible for mapping multilingual input representations into either a shared language-agnostic space \citep{chen2023bigger,starace2023probing,wang2024sharing,chen2024journey,wendler-etal-2024-llamas} that different languages share the common knowledge or distinct language-specific spaces \citep{tang-etal-2024-language,kojima2024multilingual,zhang-etal-2024-unveiling-linguistic} that are crucial for the accurate expression for specific languages. These findings support our assumption that LLMs naturally exhibit such separable structures, and our work leverages this inductive bias to improve multilingual performance.

Building upon such two theoretical foundations, particularly from linguistic theory, most previous works regarding multilingual enhancement have focused on aligning representations in the language-agnostic space \citep{hu2024large,berend2020massively,cao2020multilingual,karthikeyan2020cross,alaux2019unsupervised,wang2019multilingual} or aligning gradients during optimization \citep{lee2022sequential,wang2021gradient} to leverage shared features across languages. However, few works in multilingual machine translation have considered language-specific characteristics, primarily to implement routing mechanisms or modular designs to improve performance \citep{zhao2024sparse,zhang2021share}.

In contrast, our proposed \textsc{Lens} goes a step further that it leverages both language-agnostic and language-specific subspaces to comprehensively enhance multilingual performance both inheriting the theoretical soundness and demonstrating practical utility of our approach.

\end{document}

%% file: algo/ours.tex

\SetKwComment{Comment}{/* }{ */}
\SetKwInput{KwData}{In}
\SetKwInput{KwResult}{Out}

\begin{algorithm}[t]
\caption{Language Subspace Probing}\label{alg:ours}
\KwData{languages' mean embeddings $\boldsymbol{M}$, rank of subspace $r$}
\KwResult{language-agnostic subspace $\boldsymbol{M}_a$, language-specific subspace $\boldsymbol{M}_s$, coordinates $\boldsymbol{\Gamma}$}
\Comment{1) Approximate $\boldsymbol{M}$ in low rank}
$\boldsymbol{M}_a^\prime \gets \frac{1}{d} \boldsymbol{M} \boldsymbol{\mathbbm{1}}$\label{line:1}\;
$\boldsymbol{M}_{s}^\prime, \text{\_}, \boldsymbol{\Gamma}^\prime \gets \text{Top-} r \text{ SVD}\left(\boldsymbol{M}-\boldsymbol{M}_a^\prime \boldsymbol{\mathbbm{1}}^{\top}\right)$\;
$\boldsymbol{M}^\prime \gets \boldsymbol{M}_a^\prime \boldsymbol{\mathbbm{1}}^{\top}+\boldsymbol{M}_{s}^\prime {\boldsymbol{\Gamma}^\prime}^{\top}$\label{line:3}\;
\Comment{2) Force orthogonality}
$\boldsymbol{M}_a \gets \frac{1}{\|{\boldsymbol{M}^\prime}^{+} \boldsymbol{\mathbbm{1}}\|^2} {\boldsymbol{M}^\prime}^{+} \boldsymbol{\mathbbm{1}}$\label{line:4}\;
$\boldsymbol{M}_{s}, \text{\_}, \boldsymbol{\Gamma} \gets \text{Top-} r \text{ SVD}\left(\boldsymbol{M}^\prime-\boldsymbol{M}_a \boldsymbol{\mathbbm{1}}^{\top}\right)$\label{line:5}
\end{algorithm}